\pdfoutput=1

\documentclass[11pt]{article}

\usepackage{emnlp2021}
\usepackage{graphicx}
\usepackage{bm}
\usepackage{amsmath}
\usepackage{multirow}
\usepackage{makecell}
\usepackage{supertabular}
\usepackage{times}
\usepackage{latexsym}

\usepackage[T1]{fontenc}

\usepackage[utf8]{inputenc}

\usepackage{microtype}

\title{A Unified Neural Network Model for Readability Assessment \\
with Feature Projection and Length-Balanced Loss}

\author{Wenbiao Li$^{1,2}$,
        Ziyang Wang$^{1,2}$, 
        Yunfang Wu$^{1,3}$\thanks{~~Corresponding author.} \\
    $^{1}$MOE Key Laboratory of Computational Linguistics, Peking University \\ 
    $^{2}$School of Software and Microelectronics, Peking University, Beijing, China \\
    $^{3}$School of Computer Science, Peking University, Beijing, China \\
    \texttt{\{liwb, wzy232303\}@stu.pku.edu.cn},
    \texttt{wuyf@pku.edu.cn}
    }

\begin{document}
\maketitle
\begin{abstract}
For readability assessment, traditional methods mainly employ machine learning classifiers with hundreds of linguistic features. Although the deep learning model has become the prominent approach for almost all NLP tasks, it is less explored for readability assessment.
In this paper, we propose a BERT-based model with feature projection and length-balanced loss (BERT-FP-LBL) for readability assessment. Specially, we present a new difficulty knowledge guided semi-supervised method to extract topic features to complement the traditional linguistic features. 
From the linguistic features, we employ projection filtering to extract orthogonal features to supplement BERT representations.
Furthermore, we design a new length-balanced loss to handle the greatly varying length distribution of data.
Our model achieves state-of-the-art performances on two English benchmark datasets and one dataset of Chinese textbooks, and also achieves the
near-perfect accuracy of 99\% on one English dataset. Moreover, our proposed model obtains comparable results with human experts in consistency test.

\end{abstract}

\section{Introduction}
Readability assessment is to automatically determine the difficulty level of a given text, aiming to recommend suitable reading materials to readers.
There are wide applications of readability assessment, such as automating readers’ advisory~\cite{pera2014automating}, clinical informed consent forms~\cite{perni2019assessment} and internet-based patient education materials~\cite{sare2020readability}.

Comparing with other natural language processing (NLP) tasks, readability assessment is less explored.
In the early days, researchers exploit linguistic features to develop various readability formulas, such as Flesch~\cite{flesch1948new}, Dale-Chall~\cite{dale1948formula} and SMOG~\cite{mc1969smog}. 
Later, the mainstream research~\cite{deutsch2020linguistic,hansen2021machine,lee2021pushing} is to employ machine learning models to classify a text, by designing a large number of linguistic features.
There are also works that treat it as a regression task~\cite{sheehan2010generating} or a ranking task~\cite{lee2022neural}.

Recently, unlike other NLP tasks, the introduction of deep neural networks for readability assessment does not achieve overwhelming advantages over traditional machine learning methods.
Employing neural network models for readability assessment, there are several challenges:

(1) The scale of the dataset for readability assessment is small, which restricts the performance of deep neural network models.

(2) The deep neural network model is mainly based on characters or words and the extracted features are often at a shallow level. 
As a result, words with similar functions or meanings, such as “man” and “gentleman”, are mapped into close vectors although their reading difficulties are different~\cite{jiang2018enriching}.

(3) The linguistic features designed by researchers and continuous features extracted by neural network models are from two different semantic spaces. If two kinds of features are simply concatenated, it will bring redundant information or even harmful effects to model performance.

(4) Unlike other NLP data whose length follows a normal distribution, a notable problem with the data for readability assessment is that the text length varies greatly. The texts with low difficulty are usually shorter, while texts with high difficulty are usually longer. For example, as shown in Table 1, in ChineseLR the average length of Level 1 is only 266 characters, while the average length of Level 5 is 3,299 characters.
As a result, when experimented with deep learning networks, shorter texts tend to converge much faster than those longer ones thus harm the overall performance.

In order to solve the above problems, we propose a \textbf{BERT}-based model with \textbf{F}eature \textbf{P}rojection and \textbf{L}ength-\textbf{B}alanced \textbf{L}oss (\textbf{BERT-FP-LBL}). With the pre-trained BERT as the backbone, we employ feature projection to integrate linguistic features into the neural model, and design a new length-balanced loss function to guide the training. Concretely:

\begin{itemize}

\item{We leverage BERT and a mixed-pooling mechanism to obtain text representations, which take advantage of the powerful representative ability of pre-trained model, and thus overcome the data-sparsity problem.}

\item{Beyond traditional features, we extract a set of topic features enriched with difficulty knowledge, which are high-level global features.
Specifically, based on a graded lexicon, 
we exploit a clustering algorithm to group related words belonging to the same difficulty level, which then serve as anchor words to guide the training of a semi-supervised topic model.}

\item{Rather than simple concatenation, we project linguistic features to the neural network features to obtain orthogonal features, which supplement the neural network representations.}

\item{We introduce a new length-balanced loss function to revise the standard cross entropy loss, which balances the varying length distribution of data for readability assessment.}
\end{itemize}

We conduct experiments on three English benchmark datasets, including WeeBit~\cite{vajjala2012improving}, OneStopEnglish~\cite{vajjala2018onestopenglish} and Cambridge~\cite{xia2019text}, and one Chinese dataset collected from school textbooks. Experimental results show that our proposed model outperforms the baseline model by a wide margin, and achieves new state-of-the-art results on WeeBit and Cambridge. 

We also conduct test to measure the correlation coefficient between the BERT-FP-LBL model's inference results and three human experts, and the results demonstrate that our model achieves comparable results with human experts. 
We will make our code public available, and release our Chinese feature extraction toolkit~\textbf{zhfeat}~\footnote{\url{https://github.com/liwb1219/zhfeat}}.

\section{Related Work}

\textbf{Traditional Methods.}
Early research efforts focused on various linguistic features as defined by linguists. Researchers use these features to create various formulas for readability, including Flesch~\cite{flesch1948new}, Dale-Chall~\cite{dale1948formula} and SMOG~\cite{mc1969smog}. Although the readability formula has the advantages of simplicity and objectivity, there are also some problems, such as the introduction of fewer variables during the development, and insufficient consideration of the variables at the discourse level.

\noindent\textbf{Machine Learning Classification Methods.}
\cite{schwarm2005reading} develop a method of reading level assessment that uses support vector machines (SVMs) to combine features from statistical language models (LMs), parse trees, and other traditional features used in reading level assessment. Subsequently, \cite{petersen2009machine} present expanded results for the SVM detectors.
\cite{qiu2017exploring} designe 100 factors to systematically evaluate the impact of four levels of linguistic features (shallow, POS, syntactic, discourse) on predicting text difficulty for L1 Chinese learners and further selected 22 significant features with regression.
\cite{lu2019sentence} design experiments to analyze the influence of 88 linguistic features on sentence complexity and results suggest that the linguistic features can significantly improve the predictive performance with the highest of 70.78\% distance-1 adjacent accuracy.
\cite{deutsch2020linguistic,lee2021pushing} evaluate the joint application of handcrafted linguistic features and deep neural network models. The handcrafted linguistic features are fused with the features of neural networks and fed into a machine learning model for classification.

\noindent\textbf{Neural Network Models.}
\cite{jiang2018enriching} provide the knowledge-enriched word embedding (KEWE) for readability assessment, which encodes the knowledge on reading difficulty into the representation of words.
\cite{azpiazu2019multiattentive} present a multi-attentive recurrent neural network architecture for automatic multilingual readability assessment. This architecture considers raw words as its main input, but internally captures text structure and informs its word attention process using other syntax and morphology-related datapoints, known to be of great importance to readability.
\cite{meng2020readnet} propose a new and comprehensive framework which uses a hierarchical self-attention model to analyze document readability.
\cite{qiu2021learning} form a correlation graph among features, which represent pairwise correlations between features as triplets with linguistic features as nodes and their correlations as edges.

\section{Methodology}

\begin{figure*}[htbp]
\centering
\includegraphics[width=16cm]{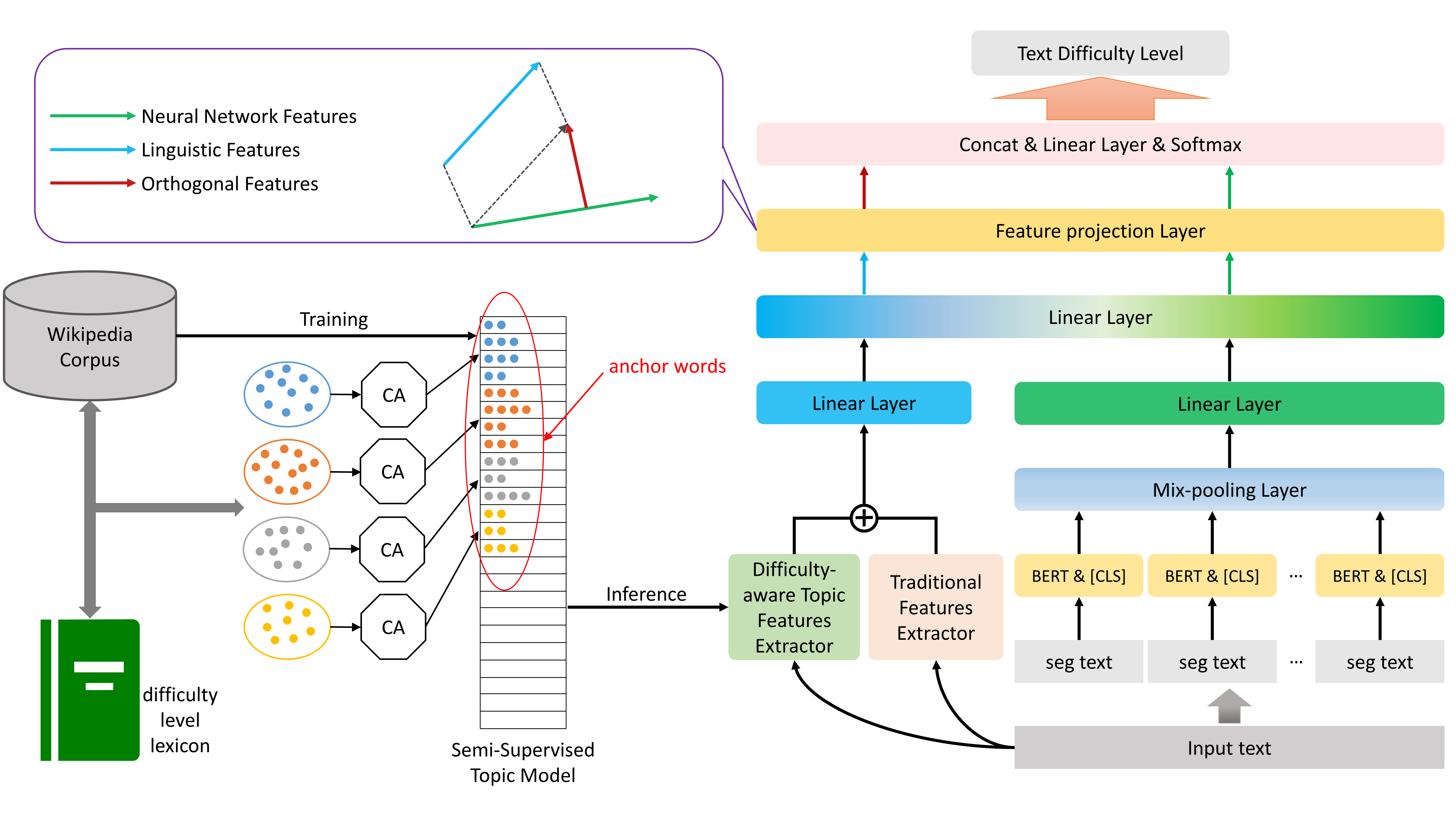}
\caption{The overall structure of our proposed model for readability assessment. \textbf{CA} represents the clustering algorithm. The input color and output color of the feature projection layer represent different types of features.}
\end{figure*}

The overall structure of our model is illustrated in Figure 1. We integrate difficulty knowledge to extract topic features using the Anchored Correlation Explanation (CorEx)~\cite{gallagher2017anchored}, and fuse linguistic features with neural network representations through projection filtering.  
Further, we propose a new length-balanced loss function to deal with the unbalanced length distribution of the readability assessment data.

\subsection{Traditional Features}
Many previous studies have proved that shallow and linguistic features are helpful for readability assessment.
For Chinese traditional features, we develop a Chinese toolkit \textbf{zhfeat} to extract character, word, sentence and paragraph features. Please refer to \textbf{Appendix A} for detailed descriptions. For English traditional features, we extract discourse, syntactic, lexical and shallow features, by directly implementing the \textbf{lingfeat}~\cite{lee2021pushing} toolkit. We denote the traditional features as $f_\alpha$.

\subsection{Topic Features with Difficulty Knowledge}
\textbf{Background.} Besides the above lexical and syntactic features, topic features provide high-level semantic information for assessing difficulty level. \cite{lee2021pushing} also leverage topic features, but they train the topic model in a purely unsupervised way without considering difficulty knowledge. Inspired by the work of Anchored Correlation Explanation (CorEx)~\cite{gallagher2017anchored}, which allows integrating domain knowledge through anchor words, we introduce word difficulty knowledge to guide the training of topic model, thus obtaining difficulty-aware topic features.

First, we introduce the concept of information bottleneck \cite{tishby2000information}, which aims to achieve a trade-off between compressing feature $X$ into representation $Y$ and preserving as much information as possible with respect to the label $Z$. Formally, the information bottleneck is expressed as:
\begin{equation}
\mathop{max}\limits_{p(y|x)} \xi I(Z:Y) - I(X:Y)
\end{equation}

\begin{equation}
\begin{aligned}
I(X_1:X_2)= & H(X_1) + H(X_2) \\  
            & -H(X1,X2)
\end{aligned}
\end{equation}
where $I(X_1:X_2)$ is the mutual information of random variables $X_1$ and $X_2$, $H(X)$ represents the entropy of the random variable $X$, and $\xi$ represents the Lagrange multiplier.

In CorEx, if we want to learn representations that are more relevant to specific keywords, we can anchor a word type $X_i$ to topic $Y_j$, and control the strength of anchoring by constraining optimization $\xi\ge1$. The optimization objective is:
\begin{equation}
\mathop{max}\limits_{\xi_{i,j},p(y_i|x)}\sum\limits_{j=1}^u\Big(\sum\limits_{i=1}^v\xi_{i,j}I(X_i:Y_j)-I(X:Y_j)\Big)
\end{equation}
where $u$ represents the number of topics, $v$ is the number of words corresponding to the topic, and $\xi_{i,j}$ represents the anchoring strength of the word $i$ to the topic $j$.

\noindent\textbf{Extracting difficulty-aware Topic Features.}
We utilize a lexicon containing words of varying difficulty levels to extract anchor words. Let $\Omega=\{L_1,L_2,..., L_k\}$ be a graded lexicon, where $L_i$ is the set of words with difficulty level $i$. $\mathcal C$ is the corpus for pre-training the topic model. First, we select out some high frequent words of each level in the corpus $\mathcal C$:

\begin{equation}
W_i=L_i \bm{\cap} \mathcal C
\end{equation}
where $\bm{\cap}$ represents the intersection operation.

For each level of words, we conduct \textbf{KMeans} clustering algorithm to do classification, and then remove isolated word categories (a single word is categorized as a class):
\begin{equation}
W_i^a={\rm{KMeans}}(W_i)
\end{equation}
The clustering result of words with the difficulty level $i$ is denoted as $W_i^a=\{\{w_{i11}^a,w_{i12}^a, ...\}, \{w_{i21}^a,w_{i22}^a, ...\}, ...\}$. Thus, the final anchor words are:
\begin{equation}
W^a=\{W_1^a, W_2^a, ..., W_k^a\}
\end{equation}

These anchor words of different difficulty levels serve as domain knowledge to guide the training of topic models:

\begin{equation}
{\rm \textbf{ATM}} = {\rm{CorEx}} (\mathcal C, anchors=W^a)
\end{equation}
where $\rm\textbf{ATM}$ represents the anchored topic model.
Then, we implement the ATM to obtain a set of topic distribution features involving difficulty information, which are denoted as $f_\beta$.

Combining traditional and topic features, we obtain the overall linguistic features:
\begin{equation}
f_\gamma = f_\alpha \oplus f_\beta
\end{equation}
where $\oplus$ represents the splicing operation.

\subsection{Feature Fusion with Projection Filtering}
\textbf{BERT Representation.} We leverage the pre-trained BERT model~\cite{devlin2018bert} to obtain sentence representation.

The length distribution of data for readability evaluation varies greatly, and texts with higher difficulty are very long, which might exceed the input limit of the model. Therefore, for an input text $S$, we segment it as $\textup{S}=(s_1, s_2,..., s_m)$. For each segment, we exploit BERT to extract its semantic representation: $H_s=(h_{s_1}, h_{s_2},..., h_{s_m})$.

Further, we adopt Mixed Pooling~\cite{yu2014mixed} to extract representations of the entire text:

\begin{equation}
\begin{aligned}
f_{\eta} = &\lambda {\rm MaxPooling}(H_{s}) + \\
&(1 - \lambda){\rm MeanPooling}(H_{s})
\end{aligned}
\label{eq:feta}
\end{equation}
where $\lambda$ is a parameter 
to balance the ratio between max pooling and mean pooling.

\noindent\textbf{Projection Filtering.} To obtain better performance, we try to combine BERT representations with linguistic features. As for the method of direct splicing, since two kinds of features come from different semantic spaces, not only will it introduce some repetitive information, 
but also it may bring contradictions between some features that will harm the performance. 
When performing feature fusion, our goal is to obtain additional orthogonal features to complement each other.
Inspired by the work~\cite{qin2020feature}, which uses two identical encoders with different optimization objectives to extract common and differentiated features.
Unlike this work, our artificial features and neural features are extracted in different ways, and our purpose is to perform feature complementation.
Since the pre-trained model captures more semantic-level features through the contextual co-occurrence relationship between large-scale corpora. This is not enough for readability tasks, and the discrimination of difficulty requires some supplementary features (difficulty, syntax, etc.).
So we consider the features extracted by BERT as primary features and linguistic features as secondary ones, and then the secondary features are projected into the primary features to obtain additional orthogonal features.

Concretely, based on the linguistic features $f_\gamma$ and BERT representation $f_\eta$, we perform dimensional transformation and project them into the same vector space:
\begin{equation}
f_{\gamma} = \text{tanh}(\text{tanh}(f_\gamma\mathbf W_1+ \mathbf b_1)\mathbf W_3+ \mathbf b_3)
\end{equation}

\begin{equation}
f_{\eta} = \text{tanh}(\text{tanh}(f_\eta\mathbf W_2+ \mathbf b_2)\mathbf W_3+ \mathbf b_3)
\end{equation}
where $\mathbf W_1$, $\mathbf W_2$ and $\mathbf W_3$ are the trainable parameters, and $\mathbf b_1$, $\mathbf b_2$ and $\mathbf b_3$ are the scalar biases. 

Next, we project the secondary features into primary ones to obtain additional orthogonal features $f_{o}$:

\begin{equation}
f_{o} = f_{\gamma} - \frac{f_{\gamma} \cdot f_{\eta}}{|f_{\eta}|^2}f_{\eta}
\end{equation}

The orthogonal features are further added to the BERT representation to constitute the final text representation:
\begin{equation}
f_{\tau} = f_{o} \oplus f_\eta
\end{equation}

Finally, we compute the probability that a text belongs to the $i-th$ category by:
\begin{equation}
p_i = {\rm Softmax}(f_{\tau}\mathbf W_4 + \mathbf b_4)
\end{equation}
where $\mathbf W_4$ is the trainable parameters, and $\mathbf b_4$ are scalar biases.

\subsection{Length Balanced Loss Function}
The text length is an important aspect for determining the reading difficulty level. As shown in Table 1, a text with high difficulty level generally contains more tokens than that of low level. For example, on Cambridge dataset, the average length of Level 1 is 141 tokens, while the average length of Level 5 is 751 tokens.
When experimented with deep learning methods, texts with short length tend to converge much faster than the texts with long length that influences the final performance.

To address this issue, we revise the loss to handle varying length.
Specially, we measure the length distribution by weighting different length attributes, including the average, median, minimum and maximum length: 

\begin{equation}
\theta_i = \sum\limits_{j=1}^4 \pi_{ij},i=1,2,...,N
\end{equation}
where $\theta_i$ represents the length value of the text category $i$, $\pi_{i,1}$ ,$\pi_{i,2}$, $\pi_{i,3}$ and $\pi_{i,4}$ represent the average, median, minimum and maximum length of the $i-th$ text, respectively. $N$ is the total number of categories.

We normalize the length value to obtain the length coefficient for each category:
\begin{equation}
\kappa_i = \frac{\theta_i}{\sum\limits_{i=1}^N \theta_i}
\end{equation}

Accordingly, the final loss function for a single sample is defined as:
\begin{equation}
\mathcal L = -\sum\limits_{i=1}^{N} \kappa_i^{\rho} y_i {\rm log}(p_i)
\end{equation}
where $y_i$ is the true label of text, $\rho$ is the adjustment factor of length distribution. When $\rho=0$, it is reduced to the traditional cross entropy loss.

\begin{table*}[htbp]
\centering
\small
\begin{tabular}{cccccccccc}
\hline
\textbf{Dataset} & \multicolumn{2}{c}{\textbf{WeeBit}} & \multicolumn{2}{c}{\textbf{OneStopE}} & \multicolumn{2}{c}{\textbf{Cambridge}} & \multicolumn{2}{c}{\textbf{ChineseLR}} \\
\hline
\textbf{Level} & Passages    & Avg.Length    & Passages          & Avg.Length          & Passages           & Avg.Length          & Passages & \multicolumn{1}{c}{Avg.Length} \\
\hline
1 & 625 & 152 & 189 & 535 & 60 & 141 & 814 & 266 \\
2 & 625 & 189 & 189 & 678 & 60 & 271 & 1063 & 679 \\
3 & 625 & 295 & 189 & 825 & 60 & 617 & 1104 & 1140 \\
4 & 625 & 242 & 0  & 0  & 60 & 763 & 762 & 2165 \\
5 & 625 & 347 & 0  & 0  & 60 & 751 & 417 & 3299 \\
All & 3125 & 245 & 567 & 679 & 300 & 509 & 4160 & 1255 \\
\hline
\end{tabular}
\caption{Statistics of datasets for readability assessment. Avg.Length means the average tokens per passage.}
\end{table*}

\section{Experimental Setup}
\subsection{Datasets}
To demonstrate the effectiveness of our proposed method, we conduct experiments on three English datasets and one Chinese dataset. We split the train, valid and test data according to the ratio of 8:1:1. The statistic distribution of datasets can be found in Table 1.

\textbf{WeeBit}~\cite{vajjala2012improving} is often considered as the benchmark data for English readability assessment. It was originally created as an extension of the well-known Weekly Reader corpus. 
We downsample to 625 passages per class.

\textbf{OneStopEnglish}~\cite{vajjala2018onestopenglish} is an aligned channel corpus developed for readability assessment and simplification research. Each text is paraphrased into three versions.

\textbf{Cambridge}~\cite{xia2019text} is a dataset consisting of reading passages from the five main suite Cambridge English Exams (KET, PET, FCE, CAE, CPE). 
We downsample to 60 passages per class.

\textbf{ChineseLR.} ChineseLR is a Chinese dataset that we collected from textbooks of middle and primary school of more than ten publishers. To suit our task, we delete poetry and traditional Chinese texts. Following the standards specified in the \textit{Chinese Curriculum Standards for Compulsory Education}, we category all texts to five difficulty levels.

\subsection{Baseline Models}

\textbf{SVM.} We employ support vector machines as the traditional machine learning classifier. The input to the model is the linguistic feature $f_\gamma$. We adopt MinMaxScaler (ranging from -1 to 1) for linguistic features and use the RBF kernel function. We use the libsvm \footnote{\url{https://www.csie.ntu.edu.tw/~cjlin/libsvm/}} framework for experiments. 

\noindent\textbf{BERT.} We utilize $f_{\eta}$ in Equation \ref{eq:feta} followed by a linear layer classifier as our BERT baseline model.

\subsection{Training and Evaluation Details}
For the selection of the difficulty level lexicon $\Omega$, on the English dataset, we use the lexicon released by ~\citet{maddela2018word}, where we only use the first 4 levels. On the Chinese dataset, we use the \textit{Compulsory Education Vocabulary}~\cite{Su2019CECV}. 
The word embedding features of English and Chinese word clustering algorithms are respectively used \cite{pennington2014glove} and \cite{song2018directional}. We use the Wikipedia corpus~\footnote{\url{https://dumps.wikimedia.org/}} for pre-training the semi-supervised topic models. 
Please refer to \textbf{Appendix B} for some other details.

We do experiments using the Pytorch~\cite{paszke2019pytorch} framework. For training, we use the AdamW optimizer, the weight decay is 0.02 and the warm-up ratio is 0.1.
The mixing pooling ratio $\lambda$ is set to 0.5.
Other specific parameter settings are shown in Table 2.

For evaluation, we calculate the accuracy, weighted F1 score, precision, recall and quadratic weighted kappa (QWK). We repeated each experiment three times and reported the average score.

\begin{table}[htbp]
\centering
\small
\begin{tabular}{cccccc}
\hline
\textbf{Dataset} & \textbf{Batch} & \textbf{MaxLen} & \textbf{Epoch} & \textbf{lr} & $\bm \rho$ \\
\hline
WeeBit    & 8 & 512   & 10 & 3e-5 & 0.8 \\
OneStopE  & 8 & 500×2 & 10 & 3e-5 & 0.4 \\
Cambridge & 8 & 500×2 & 10 & 3e-5 & 0.6 \\
ChineseLR & 2 & 500×8 & 10 & 3e-5 & 0.4 \\
\hline
\end{tabular}
\caption{Part of the hyperparameter settings, where $500 \times n$ means to split the text into $n$ segments with a length of 500 tokens.}
\end{table}

\begin{table*}[htbp]
\centering
\small
\begin{tabular}{ccccc|ccc}
\hline
\textbf{Dataset} & \textbf{Metrics} & 
\textbf{Qiu-2021} & \textbf{Mar-2021} & \textbf{Lee-2021} & 
\textbf{SVM} & \textbf{BERT} & \textbf{BERT-FP-LBL} \\
\hline
\multirow{5}{*}{WeeBit}
& Accuracy & 87.32 & 85.73 & 90.50 & 79.37 & 91.11 & \textbf{92.70} \\
& F1    & -   & 85.81 & 90.50 & 79.27 & 91.07 & \textbf{92.73} \\
& Precision & -   & 86.58 & 90.50 & 79.26 & 91.42 & \textbf{92.89} \\
& Recall  & -   & 85.73 & 90.40 & 79.37 & 91.11 & \textbf{92.70} \\
& QWK    & -   & 95.27 & 96.80 & 93.22 & 97.36 & \textbf{97.78} \\
\hline
\multirow{5}{*}{OneStopE}
& Accuracy & 86.61 & 78.72 &     99.00 & 89.47 & 97.66 & \textbf{99.42} \\
& F1    & -   & 78.88 & \textbf{99.50} & 89.32 & 97.66 & 99.41 \\
& Precision & -   & 79.77 & \textbf{99.50} & 89.41 & 97.83 & 99.44 \\
& Recall  & -   & 78.72 & \textbf{99.60} & 89.47 & 97.66 & 99.42 \\
& QWK    & -   & 82.45 & \textbf{99.60} & 92.31 & 92.98 & 98.25 \\
\hline
\multirow{5}{*}{Cambridge}
& Accuracy & 78.52 & -   & 76.30 & 83.33 & 82.22 & \textbf{87.78} \\
& F1    & -   & -   & 75.20 & 83.45 & 81.97 & \textbf{87.73} \\
& Precision & -   & -   & 79.20 & \textbf{90.91} & 82.96 & 89.46 \\
& Recall  & -   & -   & 75.30 & 83.33 & 82.22 & \textbf{87.78} \\
& QWK    & -   & -   & 91.90 & 91.97 & 94.65 & \textbf{96.87} \\
\hline
\hline
\multirow{5}{*}{ChineseLR}
& Accuracy & -   & -   & -   & 76.67 & 75.16 & \textbf{78.89} \\
& F1    & -   & -   & -   & 76.53 & 75.05 & \textbf{78.75} \\
& Precision & -   & -   & -   & 76.47 & 75.95 & \textbf{79.43} \\
& Recall  & -   & -   & -   & 76.67 & 75.16 & \textbf{78.89} \\
& QWK    & -   & -   & -   & 90.60 & 90.40 & \textbf{91.63} \\
\hline
\end{tabular}
\caption{Experimental results on both English and Chinese datasets for readability assessment. We compare our method with the recent three works, including Qiu-2021~\cite{qiu2021learning}, Mar-2021~\cite{martinc2021supervised} and Lee-2021~\cite{lee2021pushing}.}
\end{table*}

\section{Results and Analysis}

\subsection{Overall Results}
The experimental results of all models are summarized in Table 3. First of all, it should be noted that there are only a few studies on readability assessment, and there is no unified standard for data division and experimental parameter configuration. This has led to large differences in the results of different research works.

Our BERT-FP-LBL model achieves consistent improvements over the baselines on all four datasets, which validates the effectiveness of our proposed method. In terms of F1 metrics, our method improves WeeBit and ChineseLR by 1.66 and 3.7 compared to the baseline BERT model. Overall, our model achieves state-of-the-art performance on WeeBit and Cambridge. On OneStopEnglish, our model also achieves competitive results compared to previous work~\cite{lee2021pushing}, also achieving near-perfect classification accuracy of 99\%.

Comparing the experimental results of SVM and the base BERT, it can be observed that on Cambridge and ChineseLR, SVM outperforms BERT.
We believe this benefits from the linguistic features of our design.

\begin{table}[htbp]
\centering
\small
\begin{tabular}{lcc}
\hline
\makecell[c]{\textbf{Model}} & \textbf{WeeBit} & \textbf{ChineseLR} \\
\hline
BERT-FP-LBL    & 92.73 & 78.75     \\
-AW        & 92.42 & 78.23     \\ 
-TFDK       & 92.12 & 77.48     \\
-FP        & 92.25 & 78.27     \\
-LBL       & 91.76 & 76.94     \\
\hline
\end{tabular}
\caption{Ablation study in terms of F1 metric.
\textbf{$-$AW} means to remove the anchor words. \textbf{$-$TFDK} means remove the difficulty-aware topic features. \textbf{$-$FP} means that the linguistic features and neural network features are directly spliced without using projection filtering. \textbf{$-$LBL} means training using the standard cross-entropy loss function ($\rho=0$).}
\end{table}

\subsection{Ablation Study}
To illustrate the contribution of each module in our model, we conduct ablation experiments on WeeBit and ChineseLR, and the results are reported in Table 4.

When \textbf{AW} is removed, the CorEx changes from semi-supervised to unsupervised. The F1 scores of WeeBit and ChineseLR drop by 0.31 and 0.52, respectively, and when \textbf{TFDK} is removed, the corresponding F1 scores drop by 0.61 and 1.27, respectively. This indicates that our topic features incorporating difficulty knowledge indeed contribute to readability assessment.

Furthermore, when \textbf{FP} is removed, as described in Section 3.3, the simple splice operation brings some duplication or even negative information to the model. The F1 scores of WeeBit and ChineseLR both drop by 0.48.

Finally, when \textbf{LBL} is removed, the F1 scores of WeeBit and ChineseLR drop by 0.97 and 1.81, respectively. We believe that the difference in the length distribution of the dataset affects the convergence speed of different categories, which in turn will have an impact on the results. Besides, 
the drop in F1 metric is much more severe on ChineseLR than on WeeBit, and this result can be attributed to the more severe length imbalance on ChineseLR as shown in Table 1.



\begin{figure*}
\begin{minipage}[t]{0.5\linewidth}
\centering
\includegraphics[scale=0.5]{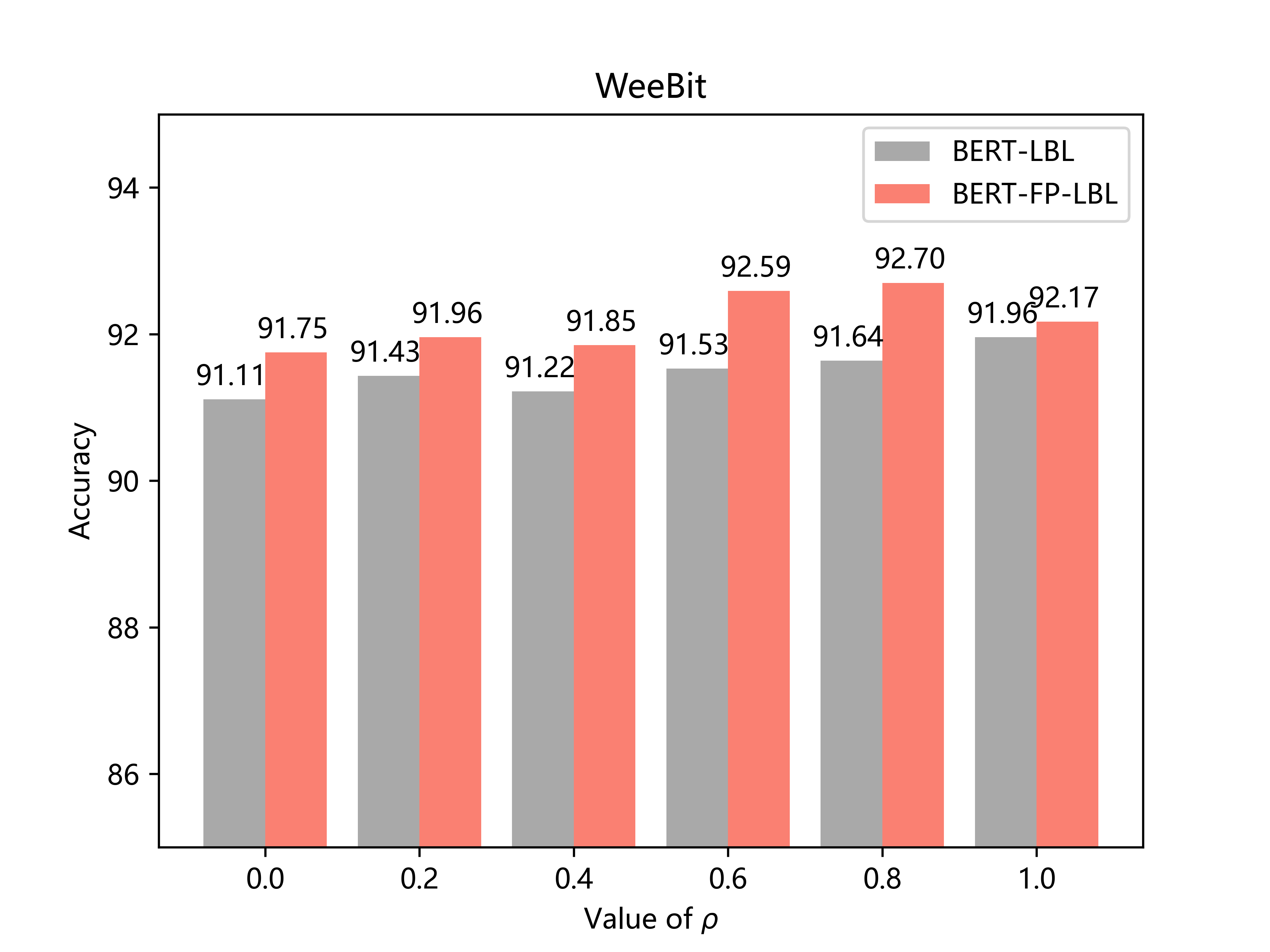}
\label{fig:side:a}
\end{minipage}%
\begin{minipage}[t]{0.5\linewidth}
\centering
\includegraphics[scale=0.5]{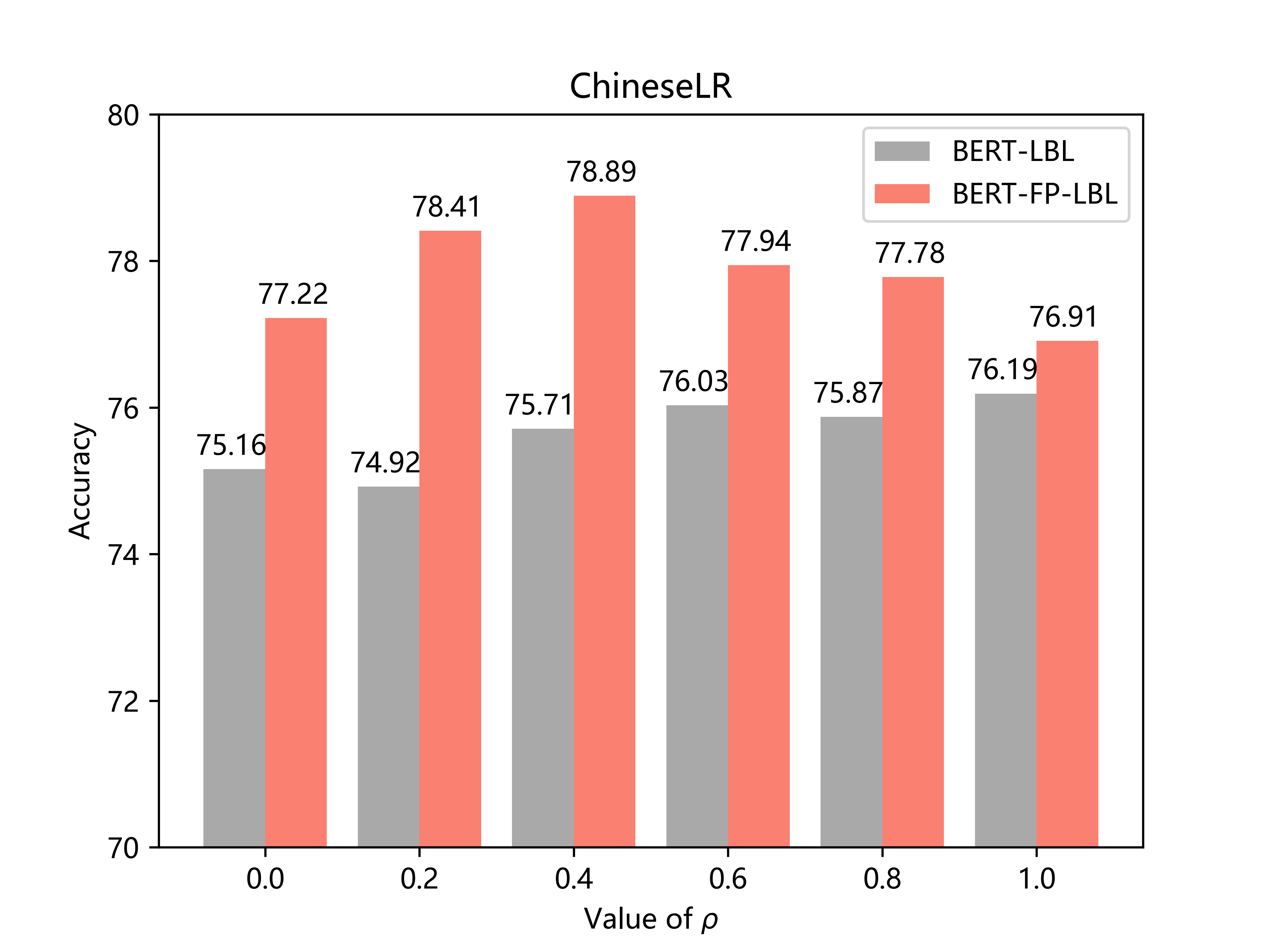}
\label{fig:side:b}
\end{minipage}
\caption{Influences of LBL on classification accuracy.}
\end{figure*}

\subsection{Analysis on the Length-balanced Loss}
To explore the effect of length-balanced loss, we set different $\rho$ to conduct experiments. The larger the $\rho$ is, the difference between the loss of different categories is bigger. The loss difference leads to different convergence rates.
When $\rho$ is 0, the loss function is the standard cross entropy loss, and there is no difference in the loss contributed by different categories. The specific results are shown in Figure 2.

For BERT, the optimal value of $\rho$ is relatively large, which means the model needs a relatively big difference in the loss to solve the problem of unbalanced text length. This indicates that there are indeed differences in the convergence speed between different classes, and this difference can be reduced by correcting the loss contributed by different classes.
After adding orthogonal features, the optimal value of $\rho$ is relatively small. We think that whether the text is short or long, the number of parameters of its corresponding orthogonal features is fixed and does not require the length-balanced loss to adjust. So, when BERT features are combined with orthogonal features, the optimal value of $\rho$ will be lower than that in BERT alone.

In addition, the optimal value of $\rho$ on WeeBit is 0.8, while the optimal value of $\rho$ on ChineseLR is 0.4. This is perhaps because the WeeBit dataset has a small span of length distribution (maximum 512 truncation), and we need to relatively amplify the differences between different categories. However, the length distribution of the ChineseLR dataset has a large span (500×8), and we need to relatively narrow the differences between different categories.

Of course, the optimal value of $\rho$ is related to the specific data distribution, which is a parameter that needs to grid search.
Generally speaking, when the length difference between different categories is small, we set $\rho$ relatively large, and when the length difference between different categories is large, we set $\rho$ relatively small.

\begin{figure*}[!t]
\begin{minipage}[t]{0.33\linewidth}
\centering
\includegraphics[scale=0.33]{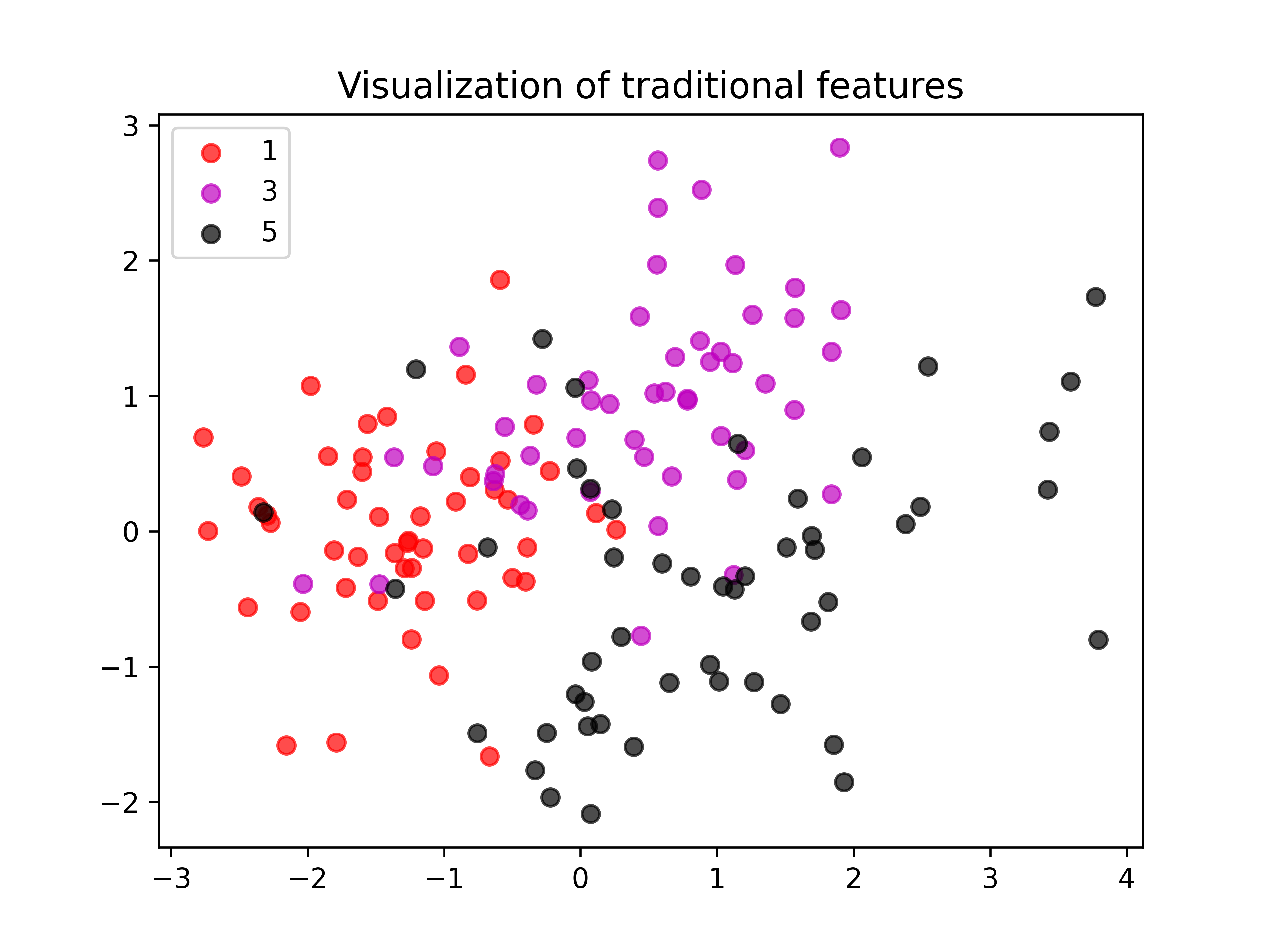}
\end{minipage}
\begin{minipage}[t]{0.33\linewidth}
\centering
\includegraphics[scale=0.33]{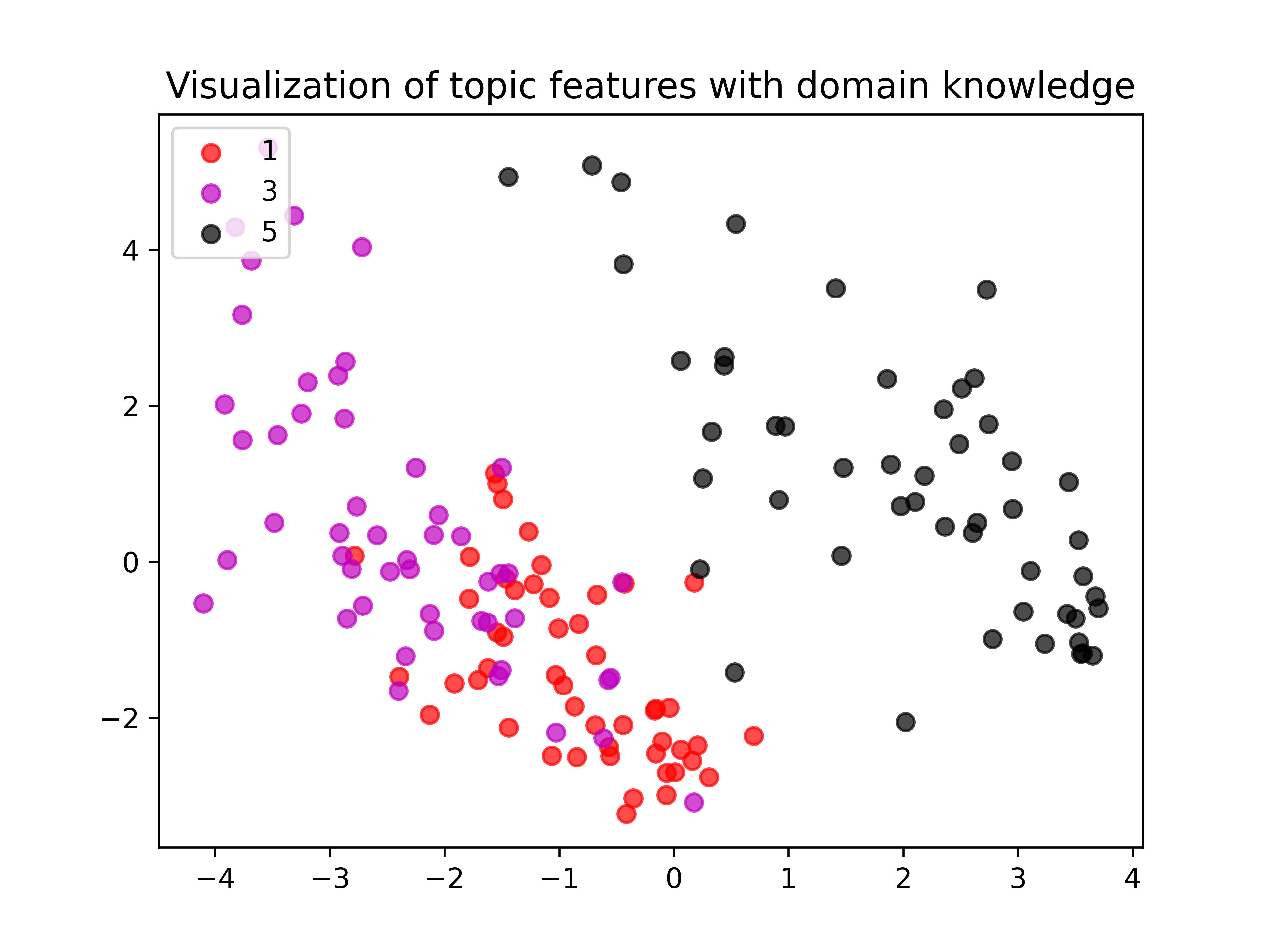}
\end{minipage}
\begin{minipage}[t]{0.33\linewidth}
\centering
\includegraphics[scale=0.33]{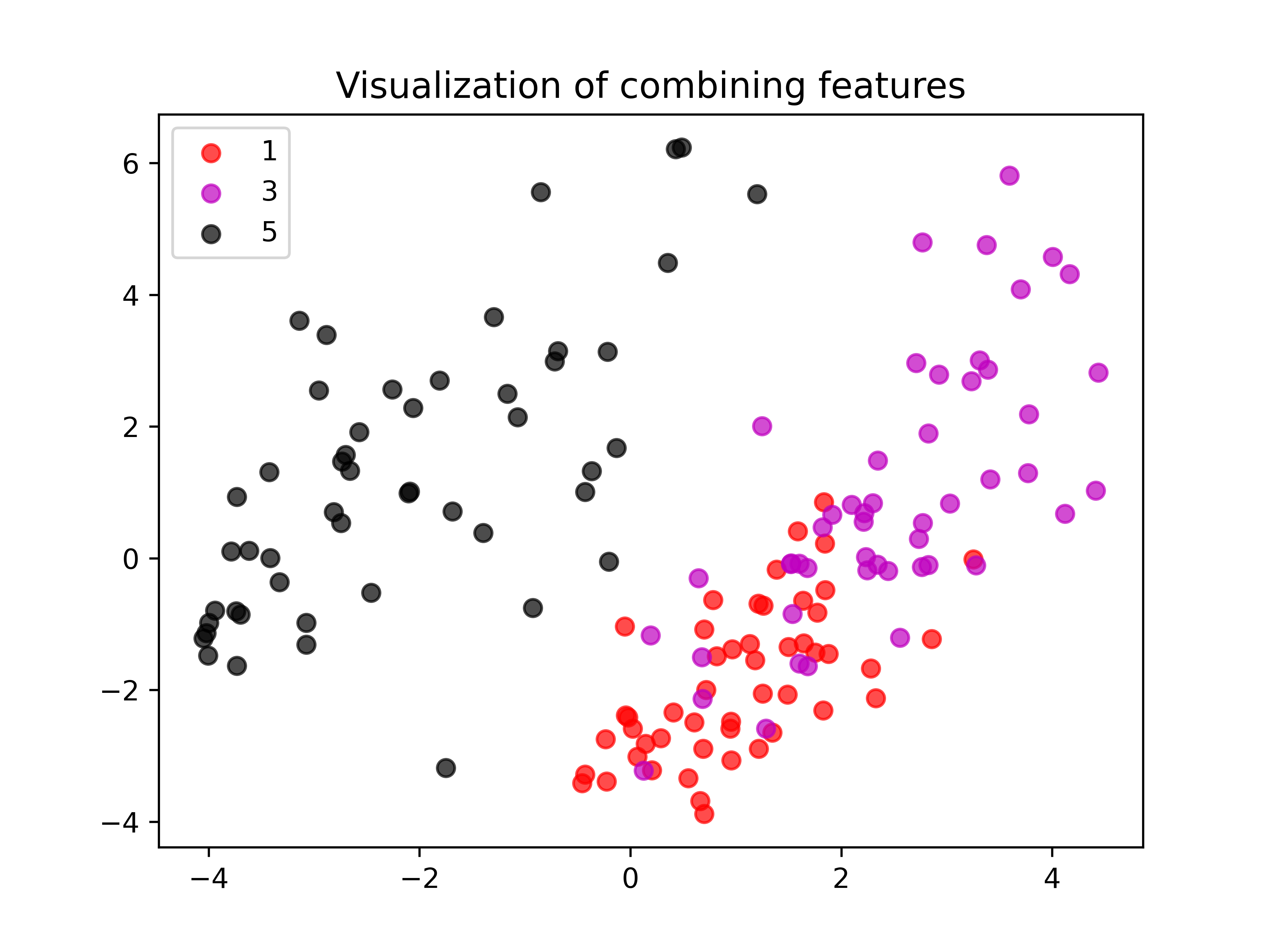}
\end{minipage}
\caption{Visualization of different kinds of features on WeeBit.}
\end{figure*}

\begin{figure*}[!t]
\begin{minipage}[t]{0.33\linewidth}
\centering
\includegraphics[scale=0.33]{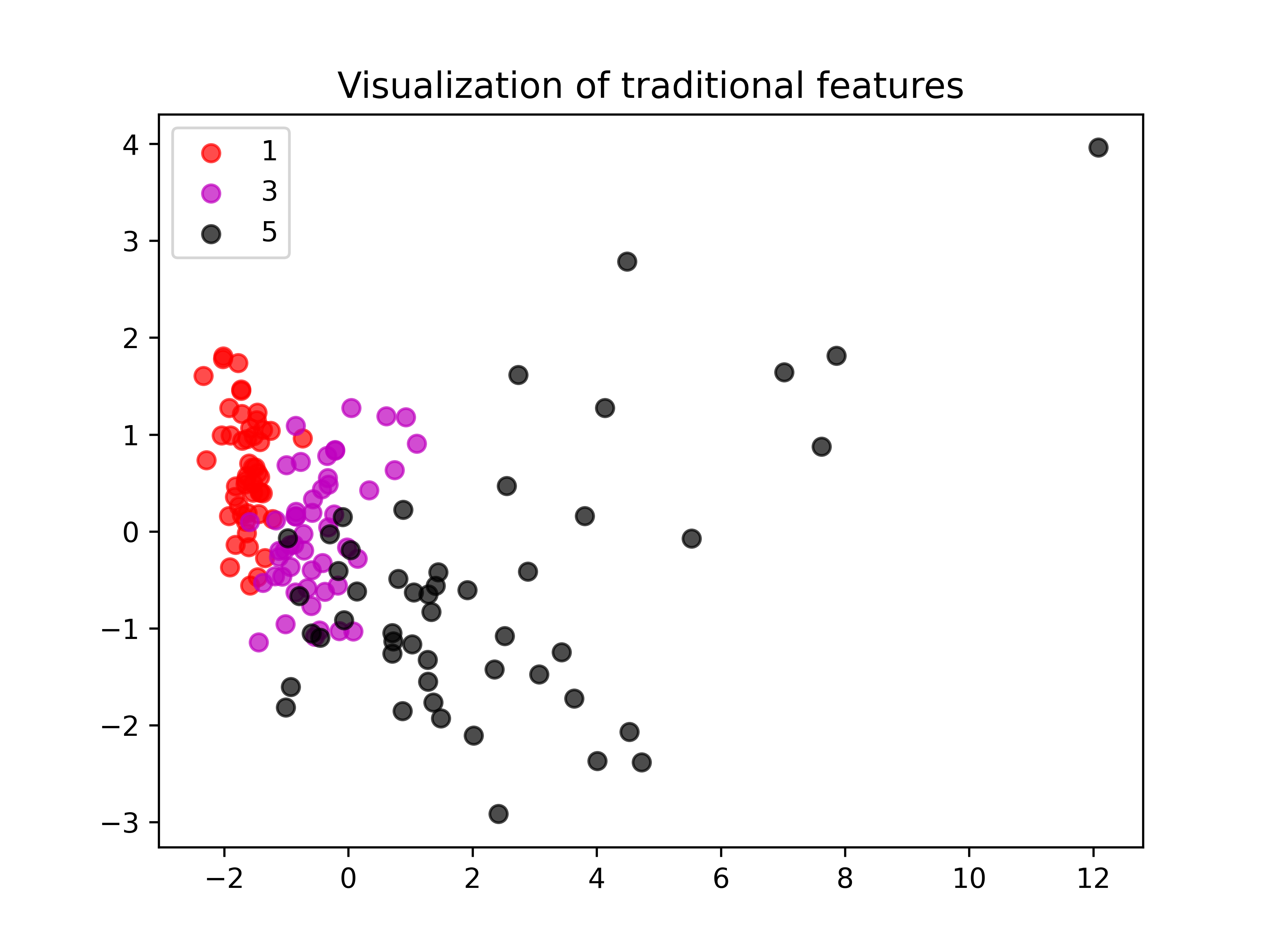}
\end{minipage}
\begin{minipage}[t]{0.33\linewidth}
\centering
\includegraphics[scale=0.33]{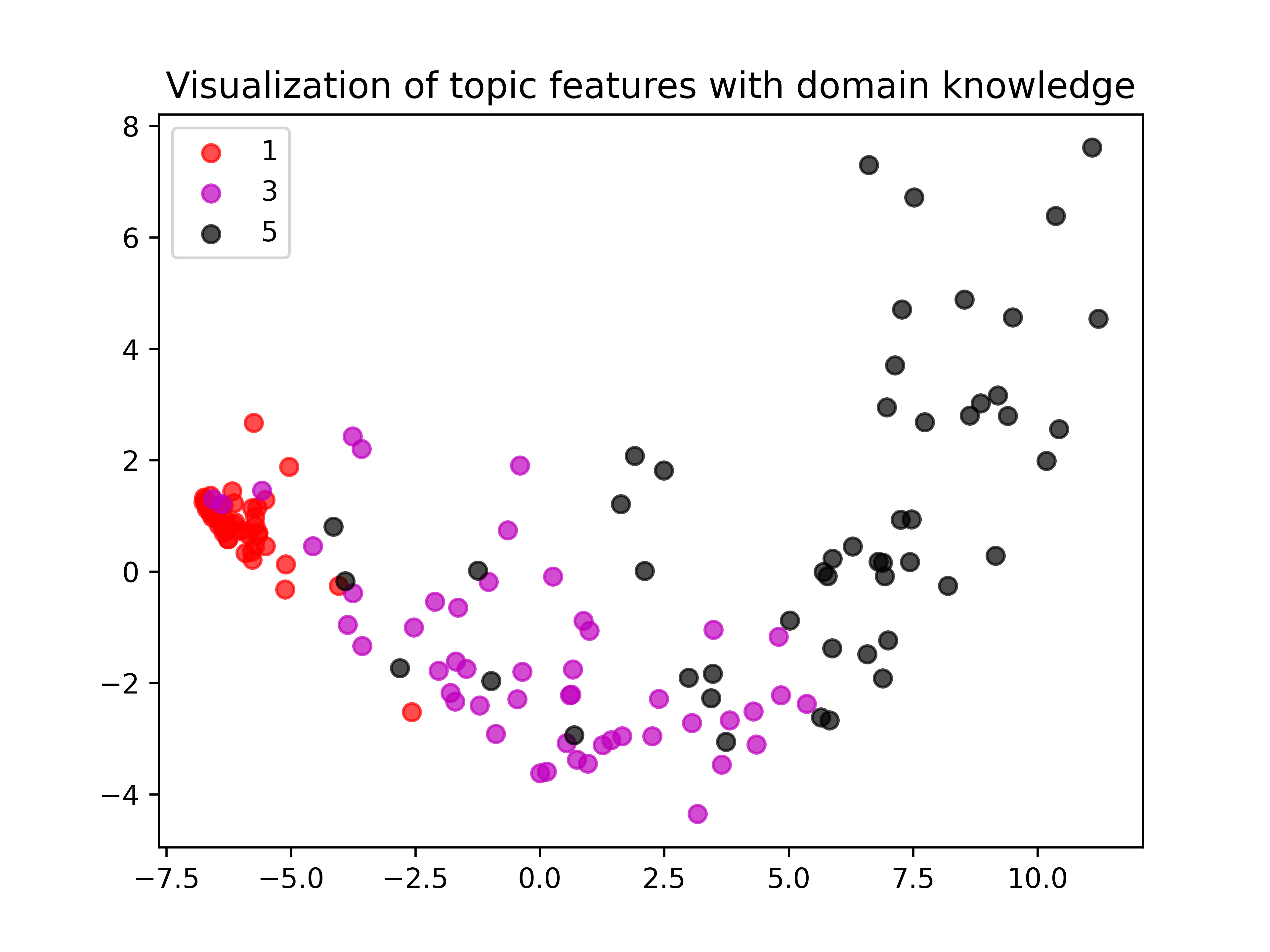}
\end{minipage}
\begin{minipage}[t]{0.33\linewidth}
\centering
\includegraphics[scale=0.33]{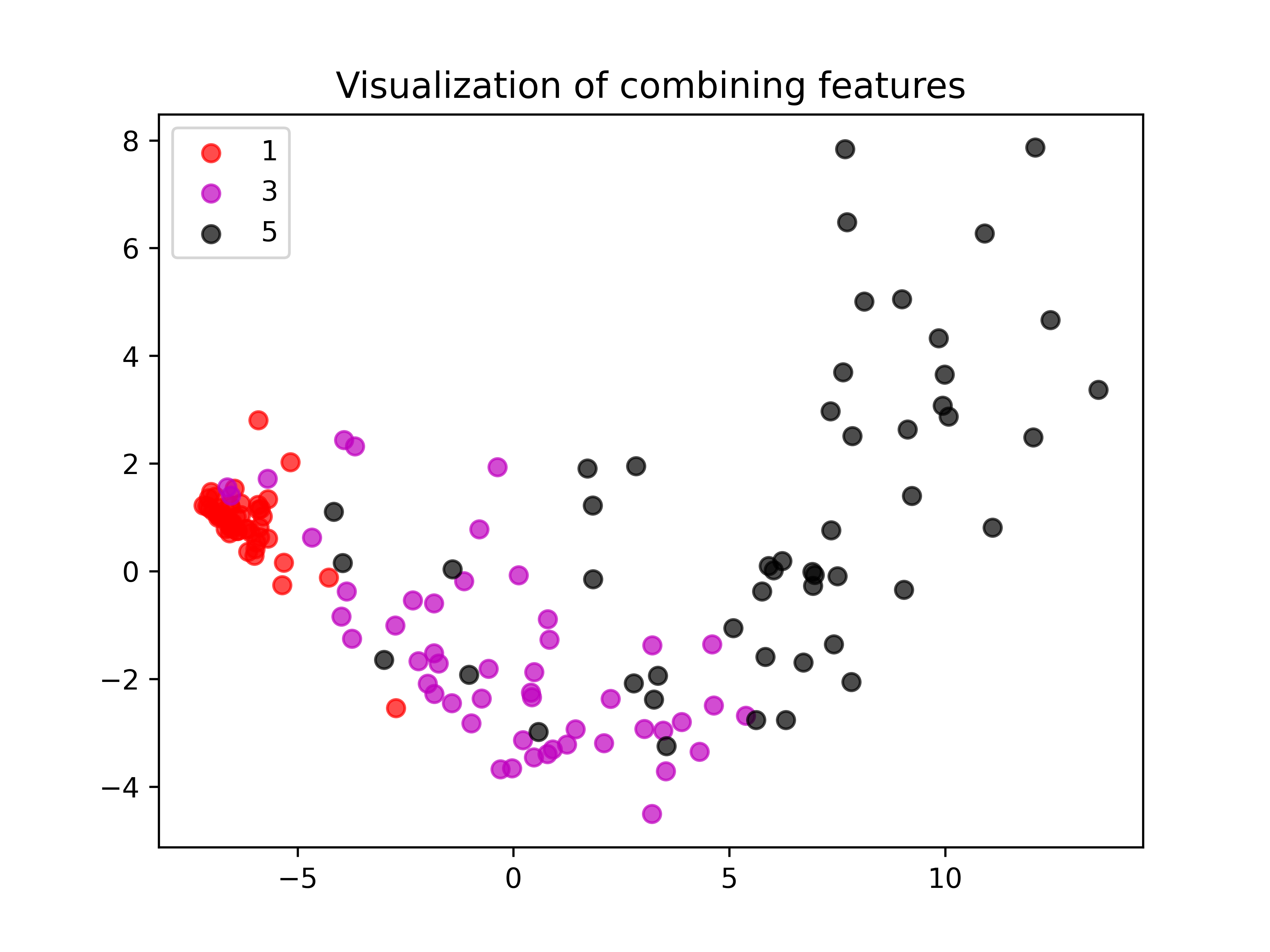}
\end{minipage}
\caption{Visualization of different kinds of features on ChineseLR.}
\end{figure*}

\subsection{Analysis on the Difficulty-aware Topic Features}
To further explore the impact of topic features with domain knowledge, we visualize the  traditional features $f_\alpha$, difficulty-aware
topic features $f_\beta$ and combining features $f_\gamma$. 
Specifically, On WeeBit and ChineseLR, we randomly selected 50 samples from each level of 1, 3 and 5 for visualization, as shown in Figure 3 and 4.

For texts of completely different difficulty, their traditional features are near in the latent space. 
This shows that traditional features pay more attention to semantic information rather than reading difficulty.
By adding difficulty-aware topic features, texts of different difficulty are better differentiated. Further, the combination of two kinds of features achieves a better ability to distinguish reading difficulty.

\subsection{Consistency Test with Human Experts}
To judge the difficulty level of a text is also a hard task for humans, and so we conduct experiments to investigate how consistent the model's inference results are with human experts. We collected 200 texts from extracurricular reading materials, and hired three elementary school teachers to do double-blind labeling. Each text is required to annotate with an unique label 1/2/3, corresponding to the first/second/third level.

Our model (denoted as M4) is regarded as a single expert that is equal to the other three human experts (E1/E2/E3). We calculate the Spearman correlation coefficient of annotation results between each pair, and report the results in Table 5.

\begin{table}[htbp]
\centering
\small
\begin{tabular}{cllll}
\hline
Rater & \makecell[c]{E1} & \makecell[c]{E2} & \makecell[c]{E3} & \makecell[c]{M4} \\
\hline
E1  & 1.000    & -     & -     & -   \\
E2  & $0.922^{\star\star}$ & 1.000   & -     & -   \\
E3  & $0.829^{\star\star}$ & $0.833^{\star\star}$ & 1.000   & -   \\
M4  & $0.836^{\star\star}$ & $0.820^{\star\star}$ & $0.807^{\star\star}$ & 1.000 \\
\hline
\end{tabular}
\caption{The Spearman correlation coefficient between four experts, where M4 is our model. $^{\star\star}$ indicates a significant correlation at the 0.01 level (two-tailed).}
\end{table}

On the whole, there is a significant correlation at the 0.01 level between human experts (E1, E2 or E3) and our model.
On the one hand, there is still a certain gap between the model and human experts (E1 and E2).
On the other hand, the inference results of our model are comparable with the human expert E3. Especially, when E1 is adopted as the reference standard, the consistency of our model prediction is slightly higher than that of E3 (0.836 vs. 0.829). When E2 is regarded as the reference standard, the consistency of our model prediction is slightly lower than that of E3.

Although there is no unified standard for the definition of "text difficulty", which relies heavily on the subject experiences of experts, our model achieves competitive results with human experts.

\section{Conclusions}
In this paper, we propose a unified neural network model \textbf{BERT-FP-LBL} for readability assessment. We extract difficulty-aware topic features through the Anchored Correlation Explanation method, and fuse linguistic features with BERT representations via projection filtering. We propose a length-balanced loss to cope with the imbalance length distribution. 
We conduct extensive experiments and detailed analyses on both English and Chinese datasets. The results show that our method achieves state-of-the-art results on three datasets and near-perfect accuracy of 99\% on one English dataset.

\section*{Limitations}
From the perspective of experimental setup, there is no uniform standard for data division and experimental parameter configuration due to less research on readability assessment. This leads to large differences in the results of different studies~\cite{qiu2021learning,martinc2021supervised,lee2021pushing}, and the results of the corresponding experiments are not comparable. Therefore, objectively speaking, our comparison object is only the baseline model, which lacks a fair comparison with previous work.

From the perspective of readability assessment task, since different datasets have different difficulty scales and different length distributions. In order to ensure the performance on the dataset as much as possible, our length-balanced loss parameters are mainly calculated according to the length distribution of the corresponding dataset, and it is impossible to transfer across datasets directly, which is also a major difficulty in this field. In cross-dataset and cross-language scenarios, there is a lack of a unified approach. Without new ways to deal with the difficulty scales of different datasets, or without large public datasets, developing a general readability assessment model will always be challenging.

\section*{Acknowledgement}
This work is supported by the  National Natural Science  Foundation of China  (62076008), the Key Project of Natural Science Foundation of China (61936012) and the National Hi-Tech RD Program of China (No.2020AAA0106600).

\bibliography{anthology,custom}
\bibliographystyle{acl_natbib}

\appendix
\clearpage
\section{Chinese Traditional Features}
\label{sec:appendix}

\begin{center}
\small
\setlength{\tabcolsep}{4pt}
\tablefirsthead{\hline\hline \textbf{Idx} & \textbf{Dim} & \textbf{Feature description} \\ \hline}
\tablehead{\hline\hline \textbf{Idx} & \textbf{Dim} & \textbf{Feature description} \\ \hline}
\tabletail{\hline\hline}
\tablelasttail{\hline\hline}
\bottomcaption{Character features description.}
\begin{supertabular}{p{0.5cm}|m{0.5cm}|p{6cm}}
1 & 1 & Total number of characters \\\hline
2 & 1 & Number of character types \\\hline
3 & 1 & Type Token Ratio (TTR) \\\hline
4 & 1 & Average number of strokes \\\hline
5 & 1 & Weighted average number of strokes \\\hline
6 & 25 & Number of characters with different strokes \\\hline
7 & 25 & Proportion of characters with different strokes \\\hline
8 & 1 & Average character frequency \\\hline
9 & 1 & Weighted average character frequency \\\hline
10 & 1 & Number of single characters \\\hline
11 & 1 & Proportion of single characters \\\hline
12 & 1 & Number of common characters \\\hline
13 & 1 & Proportion of common characters \\\hline
14 & 1 & Number of unregistered characters \\\hline
15 & 1 & Proportion of unregistered characters \\\hline
16 & 1 & Number of first-level characters \\\hline
17 & 1 & Proportion of first-level characters \\\hline
18 & 1 & Number of second-level characters \\\hline
19 & 1 & Proportion of second-level characters \\\hline
20 & 1 & Number of third-level characters \\\hline
21 & 1 & Proportion of third-level characters \\\hline
22 & 1 & Number of fourth-level characters \\\hline
23 & 1 & Proportion of fourth-level characters \\\hline
24 & 1 & Average character level \\
\end{supertabular}
\end{center}

\begin{center}
\small
\setlength{\tabcolsep}{4pt}
\tablefirsthead{\hline\hline \textbf{Idx} & \textbf{Dim} & \textbf{Feature description} \\ \hline}
\tablehead{\hline\hline \textbf{Idx} & \textbf{Dim} & \textbf{Feature description} \\ \hline}
\tabletail{\hline\hline}
\tablelasttail{\hline\hline}
\bottomcaption{Word features description.}
\begin{supertabular}{p{0.5cm}|m{0.5cm}|p{6cm}}
1 & 1 & Total number of words \\\hline
2 & 1 & Number of word types \\\hline
3 & 1 & Type Token Ratio (TTR) \\\hline
4 & 1 & Average word length \\\hline
5 & 1 & Weighted average word length \\\hline
6 & 1 & Average word frequency \\\hline
7 & 1 & Weighted average word frequency \\\hline
8 & 1 & Number of single-character words \\\hline
9 & 1 & Proportion of single-character words \\\hline
10 & 1 & Number of two-character words \\\hline
11 & 1 & Proportion of two-character words \\\hline
12 & 1 & Number of three-character words \\\hline
13 & 1 & Proportion of three-character words \\\hline
14 & 1 & Number of four-character words \\\hline
15 & 1 & Proportion of four-character words \\\hline
16 & 1 & Number of multi-character words \\\hline
17 & 1 & Proportion of multi-character words \\\hline
18 & 1 & Number of idioms \\\hline
19 & 1 & Number of single words \\\hline
20 & 1 & Proportion of single words \\\hline
21 & 1 & Number of unregistered words \\\hline
22 & 1 & Proportion of unregistered words \\\hline
23 & 1 & Number of first-level words \\\hline
24 & 1 & Proportion of first-level words \\\hline
25 & 1 & Number of second-level words \\\hline
26 & 1 & Proportion of second-level words \\\hline
27 & 1 & Number of third-level words \\\hline
28 & 1 & Proportion of third-level words \\\hline
29 & 1 & Number of fourth-level words \\\hline
30 & 1 & Proportion of fourth-level words \\\hline
31 & 1 & Average word level \\\hline
32 & 57 & Number of words with different POS \\\hline
33 & 57 & Proportion of words with different POS \\
\end{supertabular}
\end{center}

\begin{center}
\small
\setlength{\tabcolsep}{4pt}
\tablefirsthead{\hline\hline \textbf{Idx} & \textbf{Dim} & \textbf{Feature description} \\ \hline}
\tablehead{\hline\hline \textbf{Idx} & \textbf{Dim} & \textbf{Feature description} \\ \hline}
\tabletail{\hline\hline}
\tablelasttail{\hline\hline}
\bottomcaption{Sentence features description.}
\begin{supertabular}{p{0.5cm}|m{0.5cm}|p{6cm}}
1 & 1 & Total number of sentences \\\hline
2 & 1 & Average characters in a sentence \\\hline
3 & 1 & Average words in a sentence \\\hline
4 & 1 & Maximum characters in a sentence \\\hline
5 & 1 & Maximum words in a sentence \\\hline
6 & 1 & Number of clauses \\\hline
7 & 1 & Average characters in a clause \\\hline
8 & 1 & Average words in a clause \\\hline
9 & 1 & Maximum characters in a clause \\\hline
10 & 1 & Maximum words in a clause \\\hline
11 & 30 & Sentence length distribution \\\hline
12 & 1 & Average syntax tree height \\\hline
13 & 1 & Maximum syntax tree height \\\hline
14 & 1 & Syntax tree height <= 5 ratio \\\hline
15 & 1 & Syntax tree height <= 10 ratio \\\hline
16 & 1 & Syntax tree height <= 15 ratio \\\hline
17 & 1 & Syntax tree height >= 16 ratio \\\hline
18 & 14 & Dependency distribution \\
\end{supertabular}
\end{center}

\begin{center}
\small
\setlength{\tabcolsep}{4pt}
\tablefirsthead{\hline\hline \textbf{Idx} & \textbf{Dim} & \textbf{Feature description} \\ \hline}
\tablehead{\hline\hline \textbf{Idx} & \textbf{Dim} & \textbf{Feature description} \\ \hline}
\tabletail{\hline\hline}
\tablelasttail{\hline\hline}
\bottomcaption{Sentence features description.}
\begin{supertabular}{p{0.5cm}|m{0.5cm}|p{6cm}}
1 & 1 & Total number of paragraphs \\\hline
2 & 1 & Average characters in a paragraph \\\hline
3 & 1 & Average words in a paragraph \\\hline
4 & 1 & Maximum characters in a paragraph \\\hline
5 & 1 & Maximum words in a paragraph \\
\end{supertabular}
\end{center}

\section{Semi-supervised Topic Model Related Parameters}

\begin{center}
\small
\tablefirsthead{\hline\hline \textbf{Pre-training} & \textbf{English} & \textbf{Chinese}\\ \hline}
\tablehead{\hline\hline \textbf{Pre-training} & \textbf{English} & \textbf{Chinese} \\ \hline}
\tabletail{\hline\hline}
\tablelasttail{\hline\hline}
\bottomcaption{Details for pre-training the topic model.}
\begin{supertabular}{p{4.1cm}|cc}
Length range          & 300$\sim$1000  & 500$\sim$5000 \\
\hline
Items              & 209018  & 180977 \\
\hline
Topics             & 120    & 160 \\
\hline
Anchor topics          & 60    & 80 \\
\hline
Anchor strength         & 4 & 5 \\
\hline
First-level word anchor topics & 16    & 15 \\
\hline
Second-level word anchor topics & 21    & 36 \\
\hline
Third-level word anchor topics & 14    & 18 \\
\hline
Fourth-level word anchor topics & 9    & 11 \\
\end{supertabular}
\end{center}

\section{SVM model Related Hyperparameters}
\begin{center}
\small
\tablefirsthead{\hline\hline \textbf{Dataset} & \textbf{c} & \textbf{g} \\ \hline}
\tablehead{\hline\hline \textbf{Dataset} & \textbf{c} & \textbf{g} \\ \hline}
\tabletail{\hline\hline}
\tablelasttail{\hline\hline}
\bottomcaption{SVM best parameters.}
\begin{supertabular}{p{3cm}|cc}
WeeBit          & 32    & 0.004 \\
\hline
OneStopE        & 8     & 0.002 \\
\hline
Cambridge       & 16    & 0.004 \\
\hline
ChineseLR       & 64    & 0.032 \\
\end{supertabular}
\end{center}
The search range of parameter $c$ is [2, 4, 8, 16, 32, 64, 128, 256, 512, 1024, 2048, 4096, 8192, 16384, 32768], and the search range of parameter $g$ is [0.002, 0.004, 0.008, 0.016, 0.032, 0.064, 0.128, 0.256, 0.512, 1.024, 2.048, 4.096].

\end{document}